%% file: pepperMain.tex
\documentclass[letterpaper, 10 pt, conference]{ieeeconf}  

\IEEEoverridecommandlockouts                              

\usepackage{graphicx} 
\usepackage{subfigure}
\usepackage{url}

\title{\LARGE \bf
Setting Up Pepper For Autonomous Navigation And Personalized Interaction With Users}

\author{Vittorio Perera$^{1}$, Tiago Pereira$^{1}$,  Jonathan Connell$^{2}$ and Manuela Veloso$^{1}$
\thanks{$^{1}$Carnegie Mellon University, Pittsburgh}%
\thanks{$^{2}$IBM Research, T.J. Watson Research Center, Yorktown Heights}%
}

\begin{document}

\maketitle
\thispagestyle{empty}
\pagestyle{empty}

\input{intro}
\input{ros}

\input{hri}
\input{results}
\input{conclusion}

\addtolength{\textheight}{-12cm}   

\end{document}

%% file: intro.tex
\begin{abstract}

In this paper we present our work with the Pepper robot, a service robot from SoftBank Robotics. We had two main goals in this work: improving the autonomy of this robot by increasing its awareness of the environment; and enhance the robot ability to interact with its users. 
To achieve this goals, we used ROS, a modern open-source framework for developing robotics software, to provide Pepper with state of the art  localization and navigation capabilities.  
Furthermore, we contribute an architecture for effective human interaction based on cloud services. Our architecture improves Pepper speech recognition capabilities by connecting it to the IBM Bluemix Speech Recognition service and enable the robot to recognize its user via an in-house face recognition web-service.
We show examples of our successful integration of ROS and IBM services with Pepper's own software. As a result, we were able to make Pepper move autonomously in a environment with humans and obstacles. 
We were also able to have Pepper execute spoken commands from known users as well as newly-introduced users that were enrolled in the robot list of trusted users via a multi-modal interface.

\end{abstract}

\section{INTRODUCTION}\label{sec:intro}
\begin{figure}[!ht]
	\centering
	\includegraphics[scale=0.2]{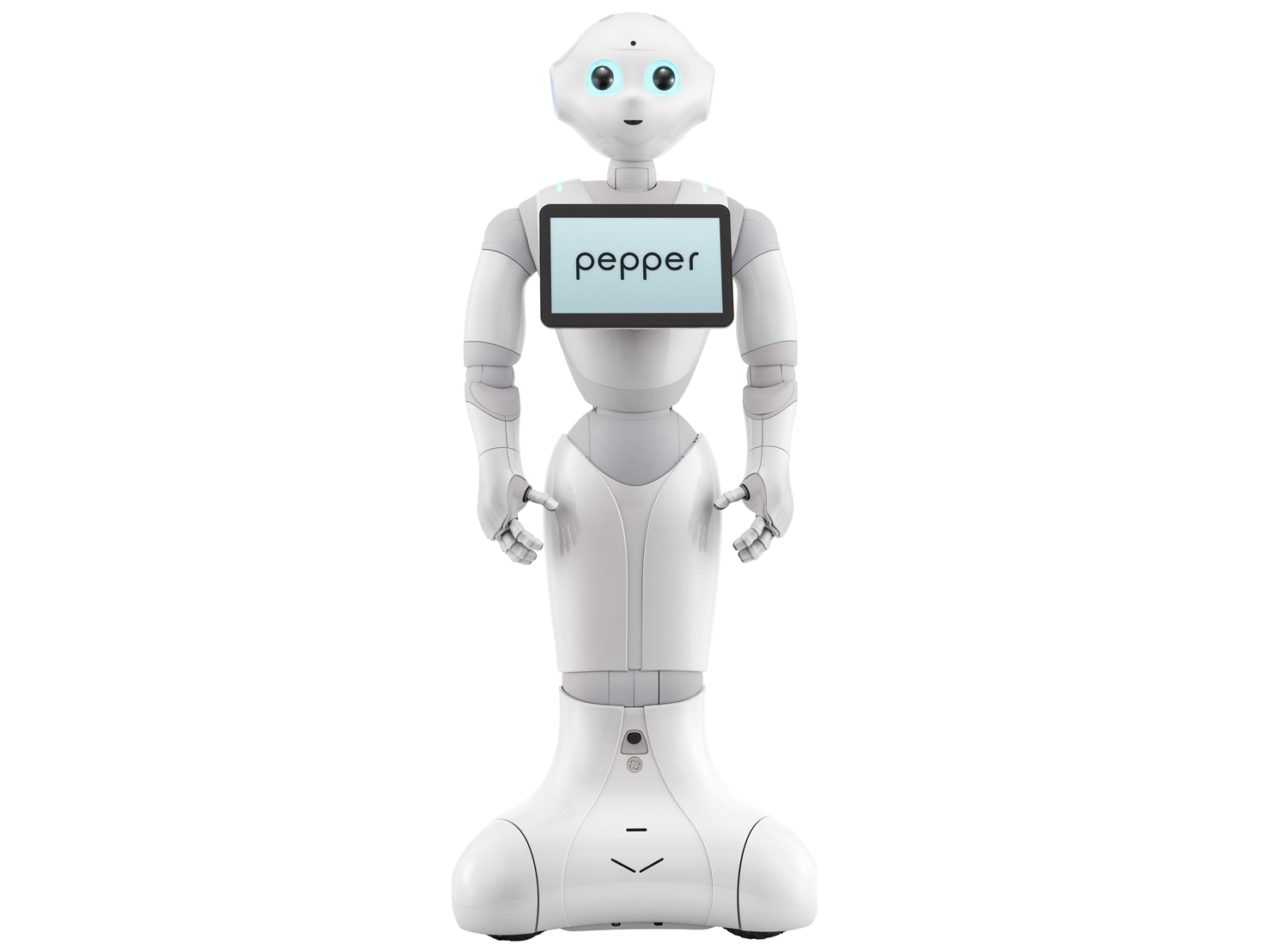}
	\caption{Pepper, the mobile human-shaped robot from SoftBank Robotics }
	\label{fig:pepper}
\end{figure}

Pepper is a service robot suited both for mobility and interaction with users, as shown in Figure~\ref{fig:pepper}. Mobility comes from the fact that this robot can move easily using its wheeled base. And its humanoid upper body, coupled with its out-of-the-box functionality such as emotion perception, allows for easy human-robot interactions.

Even though this robot has default motion, perception and interaction capabilities, we wanted to improve them in order to provide more personalized human-robot interactions. 

For that purpose, the robot needed to have a better awareness of its surroundings, from understanding the environment structure and being able to navigate safely from one place to the other, to recognizing its users and providing a clear user experience.

One possible use case is to have Pepper as a receptionist robot. It needs to be able to interact with humans through speech, and if they request information about other locations, it should give an answer either through speech, or visually using the tablet. It should also be able to escort the users partially or completely to their final destinations. Ideally, Pepper could also use any combination of these modes of interaction when acting as a reception.

Another use case is to use Pepper as a service robot in a domain where its users are known, and by being able to recognize its users, interact with them in a personalized manner, giving them opportune information. Finally, using face recognition, the robot can also learn about new users, update its knowledge base about them, and be successfully deployed in an environment continuously by adapting over time to its users.

In order to accomplish our goals for a more autonomous and personalized human-robot interaction, we contribute a software architecture that extends the proprietary NAOqi framework. We integrated ROS (Robot Operating System) with Pepper to be able to use the SLAM, localization and navigation techniques from ROS in this robot. We run those techniques locally in the robot in order to guarantee that we could keep the control loop at the desired rate.

The extended interaction capabilities, such as face recognition and speech, were designed as remote services. This design choice is explained by this services not being as time-critical as the motion control.  We integrated multiple sensors in our personalized user interaction, using touch sensors, LEDs, speech and the tablet to provide a better interaction. Microphones and cameras are used, respectively, for speech and face recognition while the LED and the tablet are used improve the ability of the robot to communicate, provide feedback to the user and make the state of the robot more transparent. In particular the eyes LEDs are used to show when the robot is listening to its use. The chest tablet use is three fold: as a caption device, displaying everything the robot says, as an input device if the user needs to type, and to display the robot state when data is being processed and an an answer cannot be provide immediately.

This paper is structured as follows: in the next section we include a brief description of the robot's hardware and software. Then we detail the software development and integration of ROS on Pepper. Then we present our solution to architecture to improve the robot capability to interact with its users. Next we demonstrate our approach with successful examples of personalized interaction and navigation autonomy. Finally, we present our conclusions and the directions for future work.

%% file: ros.tex
\section{Pepper Hardware and Software}

Pepper is a robot with a height of 1.2 meters and a total of 20 degrees of freedom, 17 in its body and 3 in its base. The base is, therefore, omnidirectional and allows for holonomic planning and control of its motion. The robot has an IMU, which coupled with the wheels' encoders, provides odometry for dead-reckoning. For obstacles avoidance, the robot has two sonars, two infrared sensors, and six lasers, three of which pointing down and other three looking for obstacles in the robot's surrounding area. It also possesses three bumpers next to the wheels, used to stop in the eventuality of a collision with obstacles.

For interaction with users, the robot offers a four element microphone array positioned on the top of the head, two loudspeakers, and three tactile sensors, one on the back of each hand and one on top of the head. Pepper also has three groups of LED that can be used for non-verbal communication. The LEDs are positioned in the eyes, on the shoulders and around loudspeaker in the ears. The tablet on Pepper's chest is an Android tablet and its possible to either develop apps that integrate with the robot or use it as a display by loading web pages, pictures or video. Both the robot and the tablet have independent wireless connectivity.

For perception, the robot has 2 cameras with a native resolution of 640*480. The two cameras are positioned, respectively, in the forehead pointing slightly upward, and the mouth pointing downward. Given the height of the robot, the top camera was the natural choice for HRI as it point toward the average height of a user. Pepper also has an ASUS Xtion 3D sensor in one of its eyes, which we used for localization and navigation.

In terms of computational power, Pepper is equipped with a 1.9GHz quad-core Atom processor and 4GB of RAM. Finally, the robot's operating system is NAOqi OS, a GNU/Linux distribution based on Gentoo. For security reasons, developers do not have root permissions, with \textit{sudo} only being available to shutdown or reboot the robot.

\section{Integrating ROS with NAOqi}

In this section we present our efforts to integrate ROS Indigo in the Pepper robot with NAOqi 2.4.3. Similar work has been done with the NAO robots~\cite{ros2nao}, but here we propose an alternative method to cross compilation~\cite{crosscomp}, using a virtual machine with Pepper's operating system, the NAOqi OS.

\subsection{NAOqi Process and Framework}

\begin{figure}[!ht]
	\centering
	\includegraphics[scale=0.5]{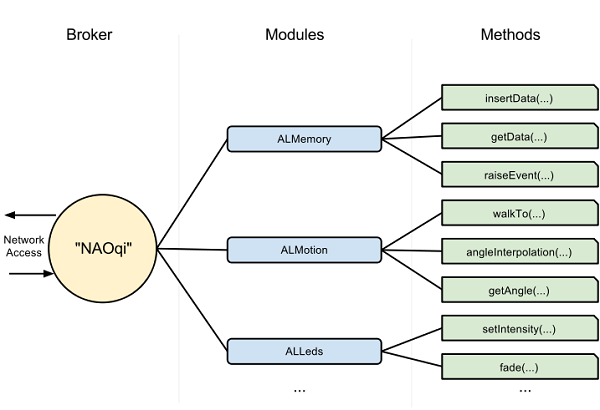}
	\caption{The NAOqi process~\cite{naoqi}}
	\label{fig:naoqi}
\end{figure}

Pepper is controlled by an executable, NAOqi, which is started automatically when the operating system NAOqi OS is started. As we show in Figure~\ref{fig:naoqi}, NAOqi works as a broker, and loads libraries that contain one or more modules that use the broker to advertise their methods.

The NAOqi process provides lookup services to find methods, and network access, allowing methods to be called and executed remotely. Local modules are in the same process, so they can share variables and call each others’ methods without serialization nor networking, allowing the fast communication. Local modules are ideal for closed loop control. Remote modules communicate using the network, and so it is impossible to do fast access using them.

The NAOqi executable comes with a list of core modules and a public API~\cite{naoqi}, with its functionality divided in groups such as
\begin{itemize}
\item \textbf{NAOqi Core}: modules that handle communications, module management, memory access, and emotion recognition;
\item \textbf{NAOqi Motion}: modules implementing animations, navigation tasks, and low level control of joints position and base velocity;
\item \textbf{NAOqi Audio}: modules controlling the animated speech, speech recognition, and audio recording;
\item \textbf{NAOqi Vision}: modules that do blob detection, photo capture, and basic localization and landmark detection;
\item \textbf{NAOqi People Perception}: modules focused in human-robot interaction, with face detection, gaze analysis and people tracking;
\item \textbf{NAOqi Sensors}: modules for reading the laser, sonar, and touch sensors 
\end{itemize}

The NAOqi Framework is a cross-platform programming framework that can be used to program the SoftBank robots. Developers can use either Python or C++ to build new modules for Pepper, which can interact with the default NAOqi modules.

Python, being an interpreted language, is very flexible and easy to run both remotely or locally in the robot. When using C++, modules can also be run locally or remotely by compiling them to the target OS. Therefore, if running new modules locally, the developer needs to cross-compile the executables for the NAOqi OS.

\subsection{ROS Interface for Localization and Navigation}

The Robot Operating System (ROS) is an open source middleware framework with libraries and tools for robot software development. It includes state-of-the-art algorithms, an inter-process communication framework and visualization tools. It is used on many robots and research groups because of its hardware abstraction and package management, with many algorithms for perception, planning, and localization.

In ROS the processing units are called nodes, which communicate either via topics or services. Topics follow the publisher/subscriber paradigm, while services work in a  client/server model. There is always a master coordination node, but other nodes can be distributed, allowing distributed operation over multiple machines.

The ROS community supports a ROS Interface node to bridge ROS and the NAOqi system~\cite{pepper-ros}. It runs both as a ROS node and a NAOqi module, and translates the NAOqi calls to ROS services and topics with standard types. 

Therefore, in our work we use ROS as a tool to enable easy integration of Pepper with state-of-the-art navigation techniques and basic visualization tools. Furthermore, once the ROS Interface is running, developers can create software for Pepper with standard types and communication, abstracting from its specific hardware. This also makes it easy to port code to and from other robots and simulations.

\subsubsection{Autonomous Navigation}

In order to navigate autonomously, the robot needs to have a map of the environment and use its odometry and perception to estimate its position on that map. The traditional approach is to use SLAM (simultaneous localization and mapping) in order to build a 2D map of the environment using a laser range finder, and then use the pre-built map to localize the robot.

Even though the robot has 3 horizontal lasers measuring the distance to its surrounding obstacles, they provide very few points, which makes it impracticable to use this sensors for SLAM or localization. The lasers on Pepper are used only to avoid colliding with obstacles. Therefore, we instead use the 3D sensor to get depth images, converting them to a simulated 2D laser scan. The converted laser scan can then be used for 2D SLAM, localization and navigation. It is also possible to rotate the head to increase the field of view of the simulated laser scan.

\begin{figure}[!ht]
	\centering
	\includegraphics[scale=0.3]{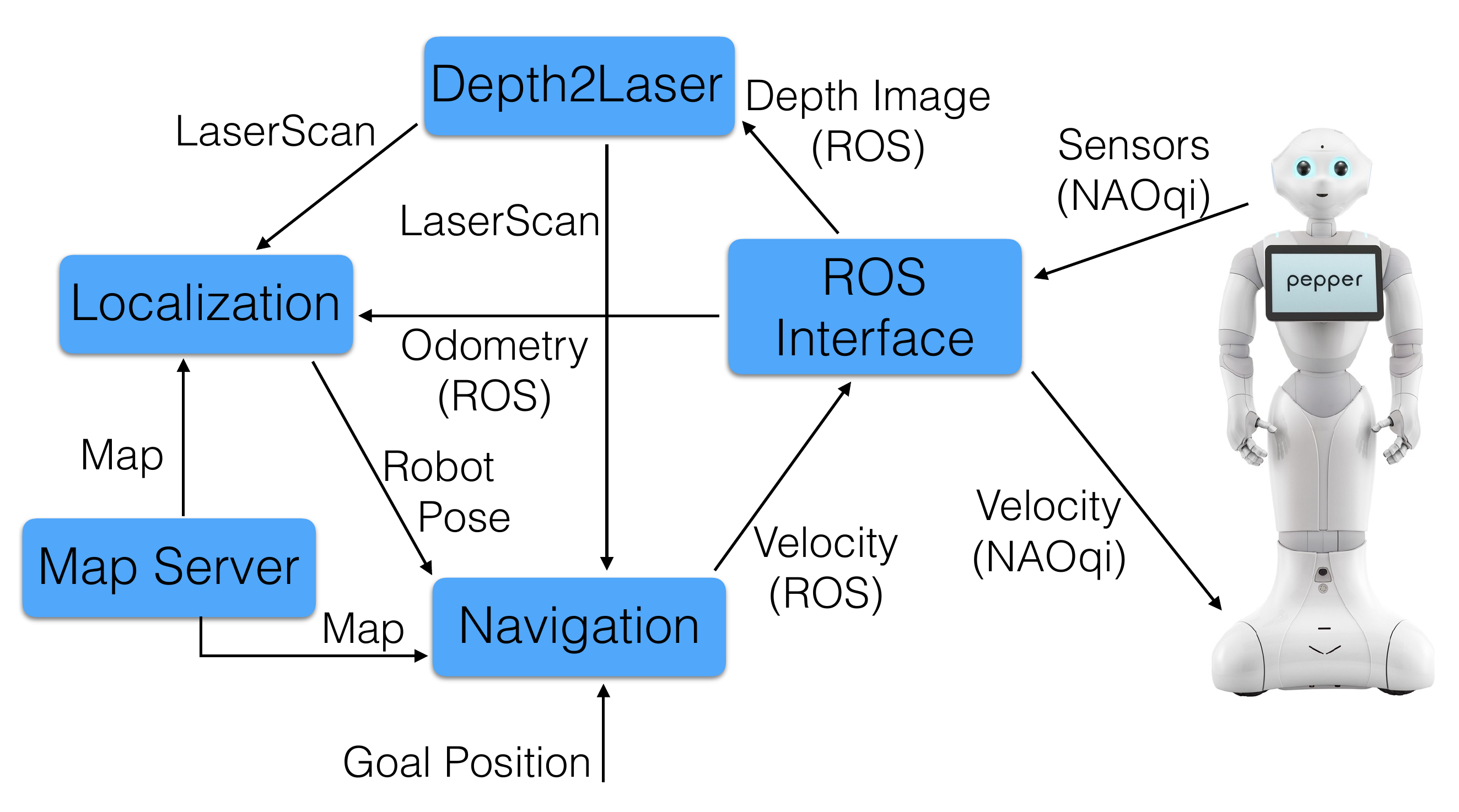}
	\caption{Software architecture for localization and navigation with ROS in the Pepper robot, with ROS nodes shown in blue, and arrows representing communication over topics.}
	\label{fig:ROS-arq}
\end{figure}

The software architecture for running localization and navigation with ROS on Pepper, presented in Figure~\ref{fig:ROS-arq}, has a central element, the ROS Interface. The ROS Interface registers as a module in the NAOqi, and then it makes calls with the NAOqi API in order to read the sensors from Pepper and send velocity commands to its base. 

The loop is closed by the remaining ROS nodes. After reading the sensor data from Pepper, the ROS interface publishes the depth image and odometry using the standard ROS types. Another node subscribes the depth image and converts it to a laser scan. The ROS community provides a simple package, \textit{depthimage\_to\_laser scan}, that can perform this transformation. However, this package is suited for images acquired from 3D sensors fixed and mounted horizontally, at a low height enabling them to easily detect obstacles in the ground. However, Pepper has its 3D sensor in the head, which can move and change orientation. Therefore, we used instead two other packages, the first converting from depth images to point clouds (\textit{depth\_image\_proc}), and the second creating the 2D laser scan from the 3D point cloud (\textit{pointcloud\_to\_laserscan}). The last node can convert the 3D data from its reference frame to other target frames, being flexible on the minimum and maximum height of points to consider for the conversion. It also allows us to do the conversion when the robot's head is tilted down. 

\subsubsection{Localization and Navigation}

With the laser scan and odometry published, and a map given by the map server, the robot is able to localize itself in the map frame using AMCL, the Adaptive Monte Carlo Localization~\cite{amcl}. Finally, using the map, the robot pose estimate from localization, and the laserscan for obstacle avoidance, the robot can autonomously navigate to a goal position. This goal position can be given from a visualization tool like rviz, or even indirectly through speech interaction as an user request. The navigation node finds the optimal path to go from the current position to the goal position, while avoiding obstacles in the map and also seen for the first time while navigating. Finally, the navigation node also finds the velocity commands that should be sent to the robot so it follows the planned path. The ROS interface then closes the loop by translating the velocity commands with standard ROS type into a function call to the NAOqi in order to send the command.

\begin{figure}[!ht]
	\centering
	\includegraphics[scale=0.35]{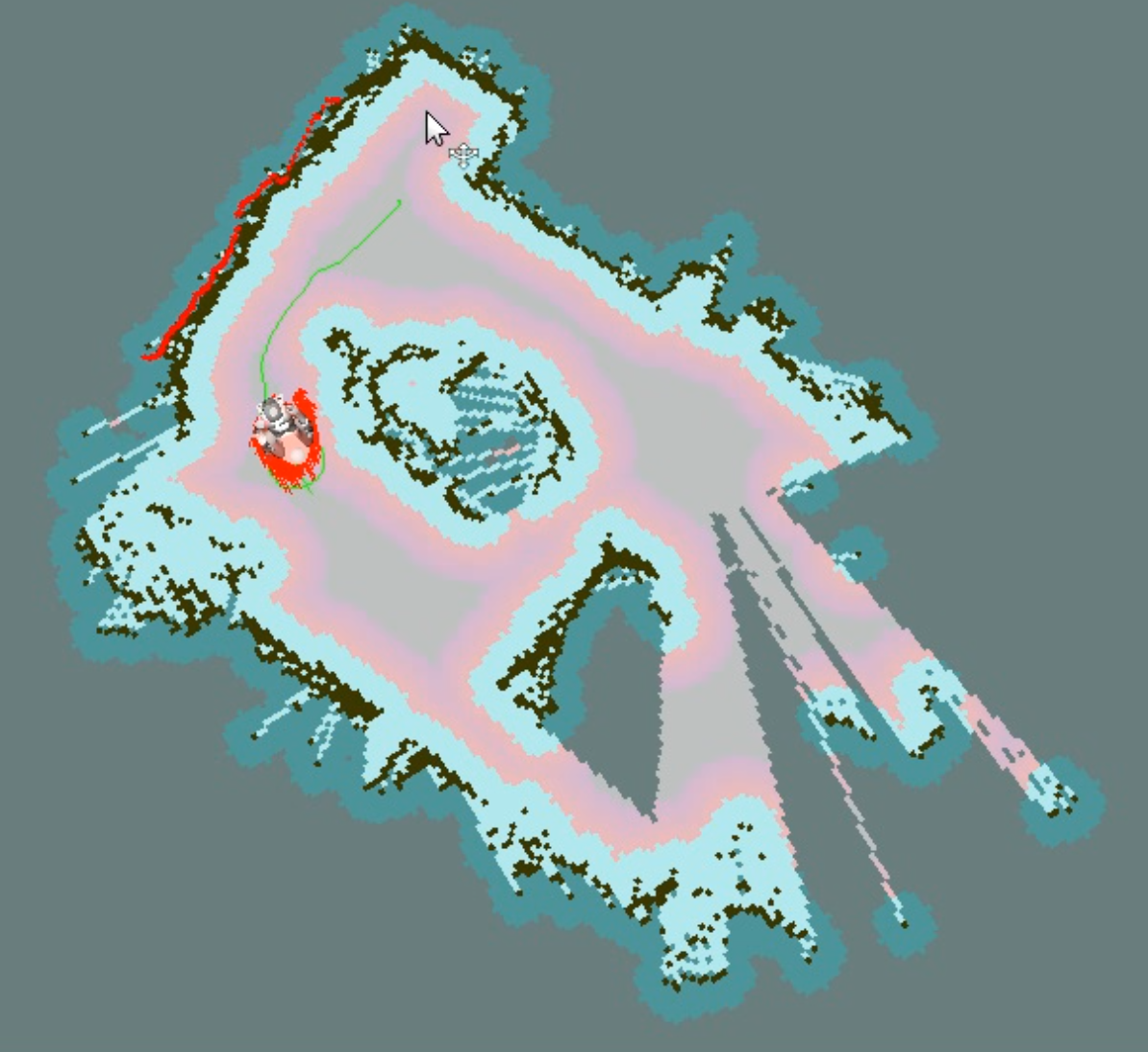}
	\caption{Visualization of a real Pepper while moving in a pre-build map of the environment: black represents obstacles, light gray is the free space, and dark grey represents the unexplored regions; the global cost map is shown in blue and light purple; the green line is the optimal path from the current position to the destination.}
	\label{fig:navigation}
\end{figure}

In Figure~\ref{fig:navigation} we show a visualization of a real Pepper localizing and navigating at a IBM Research location. The map was built with the Gmapping~\cite{gmapping} (grid-based SLAM), with black being the obstacles and light grey the free space. The laser scan simulated from the depth image is shown as red dots that overlap one wall. The global planner uses the map with inflated obstacles (ROS costmap) to determine the navigable space, and a planner finds the optimal path from the current position to the destination from that costmap. Another local costmap is determined using only the observations from the simulated laser scan. Finally, the robot position is estimated with a particle filter, represented in the image with red arrows around the robot. The size of this cloud of particles shows the uncertainty of the robot pose estimate.

\subsection{Software Development for Pepper}

Given that both ROS and NAOqi allow for a distributed architecture, the scheme of Figure~\ref{fig:ROS-arq} applies independently of where the nodes run, either locally on Pepper or remotely on other machine. However, as explained previously, anything with real-time constraints should run locally on the robot. Because localization and navigation are time-critical, we run the ROS Interface and all the other ROS nodes locally on the robot  in order to keep the control loop rate, except for the visualization, which run remotely on a Ubuntu machine.

In order to successfully implement the methodology described before, we had to install and run ROS locally on the robot. We said before that in order to run C++ on the robot we needed to cross-compile it to the robot OS. However, there is an alternative to cross-compilation, which we found easier to use for installing ROS on the robot.
 
The software development alternative is to use the virtual machine that SoftBank provides with the same OS found in Pepper. While Pepper doesn't have software development tools or package manager, the virtual machine provides developer tools such as \textit{gcc}, \textit{cmake}, and the \textit{emerge} and \textit{portage} package tree for Gentoo. More importantly, the virtual machine has root permissions, so we can compile and install any third-party system dependency needed.

The virtual machine makes it easier to build third-party libraries, which can be easily ported to the Pepper robot by copying the files through \textit{ssh}. This can be done without administration privileges on the robot, by placing the files inside some folder under the home directory and pointing the environment path variables \textit{LD\_LIBRARY\_PATH} and \textit{PYTHONPATH} to the right directories. Moreover, when installing system dependencies with the package manager \textit{emerge}, the virtual NAOqi OS will create a compressed file with header files and libraries inside a folder called \textit{opennao-distro}, under the home directory. This process makes it very easy to indirectly install system dependencies on Pepper, by installing them in the virtual machine and easily copying the files to the robot. SoftBank recommends not to upgrade the system in the virtual machine, as packages build after that may be not compatible with the NAOqi OS on the robot. 

After installing the needed system dependencies, we can compile ROS from its source-code in the virtual machine. In order to run our experiments, we needed to compile both the ROS core packages and the navigation packages as well, with all its ROS dependencies. While most of the ROS packages needed are catkin-based, some of these ROS packages needed for the navigation stack are pure cmake packages. Therefore, in order to compile those packages, we needed to create a separate workspace and compile it with the \textit{catkin\_make\_isolated} command. As we found out, this command is not completely supported on Gentoo, but that can be fixed by installing \textit{dpkg}, the package maintenance system for Debian. 

Again, we can indirectly install ROS on Pepper by compiling it in the Virtual Machine and moving the files to Pepper through \textit{ssh}, just needing to update the environment and ROS path variables to the right directories. In conclusion, these are the steps to install ROS on Pepper:
\begin{itemize}
	\item Download Virtual Machine with NAOqi OS (VM);
	\item Download ROS source-code to VM;
	\item Compile system dependencies on VM, such as \textit{log4cxx}, \textit{apr-util}, \textit{yaml}, among others;
	\item Compile ROS packages on VM with \textit{catkin\_make\_isolated} in release mode in order to compile both catkin and pure cmake packages (in order to have a successful compilation, it might be needed to separate pure cmake packages in one workspace, compiling catkin packages in another workspace with \textit{catkin\_make} instead);
	\item Copy ROS installation folder and \textit{opennao-disto} folder with system dependencies to home directory on Pepper;
	\item Change ROS variables and environment path variables \textit{LD\_LIBRARY\_PATH} and \textit{PYTHONPATH} to point the right directories.
\end{itemize}

%% file: hri.tex
\section{Effective Human Interaction On Pepper}

To enable Pepper to have natural interaction with its user we identified the need for three critical components: the ability to identify its user (i.e., face recognition), the ability to understand natural language (i.e., speech recognition) and a clear way to convey the internal state of the robot to its user. This section is going detail our design choices in the implementation of these three aspects focusing on the audio recording, the speech recognition, the face recognition and the use of the chest tablet as additional input/output device . 

\subsection{Audio Recording}
In order to be able to run any form of speech recognition the first step is to record the audio input.The NAOqi APIs allow to start and stop the recording from the head microphones with two different function calls, and the audio being recorded is immediately written on file. Although it's possible to use an attention keyword to start the speech recognition this would require that robot is always listen, and therefore recording. To avoid this we decided to start the recording through a haptic interface. The NAOqi framework natively monitors the hand touch sensor and triggers \texttt{HandTouched} event, by listening to these event we enable the users to have the robot listen by slightly pressing its arm. 

Once the robot start recording we want to have the user know that the robot is actually listening. To do so we decided to use the eyes LED, by having them blink with blue and green light the user is able to immediately tell that the robot is recording audio. 

In order to stop the recording we use a dynamically determined energy threshold over a sliding window. The NAOqi api allows to measure the current energy level of the microphones. During the first 200ms of recording we monitor these energy level and set the silence threshold $\tau_s$ to its average. The NAOqi framework offers native calls to monitor the energy level of the microphone that can be considered a measure of the noise level in the environment. Our idea is that, the user doesn't immediately start speaking after having pushed the robot hand and, therefore, we can use a small time frame to monitor what is the noise level of the environment and use it as a reference to know when the user has stopped speaking. By using a moving windows of length 1 second, shifting every 200 ms, we keep monitoring the average energy level of the recording and stop when gets close to the initial value of $\tau_s$. By doing so we enable the robot to stop the audio recording when the users finished talking; at the same time as the recording stop we also stop the eyes LED blinking to notify the user that the robot is no longer listening.

\subsection{Speech Recognition}\label{sec:speechreco}
The NAOqi framework offer some native functions for speech recognition but they require to specify a priori the vocabulary the robot will be able to recognize. In order to empower Pepper's user with any kind of language we decided instead to connect Pepper to the IBM Watson Speech to Text service~\cite{ibm-speech}. This service runs on the IBM Bluemix Cloud service and requires the audio file recorded to be sent to the remote server. 

\begin{figure}[!h]
	\centering
	
	\subfigure[]{%
		\includegraphics[width=8cm, keepaspectratio]{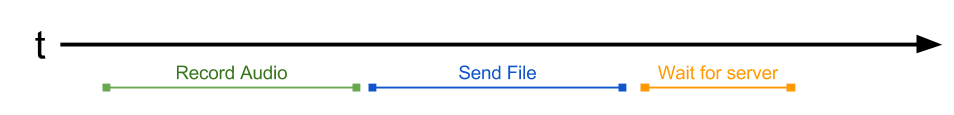}
	}
	\quad
	
	\subfigure[]{
		\includegraphics[width=8cm, keepaspectratio]{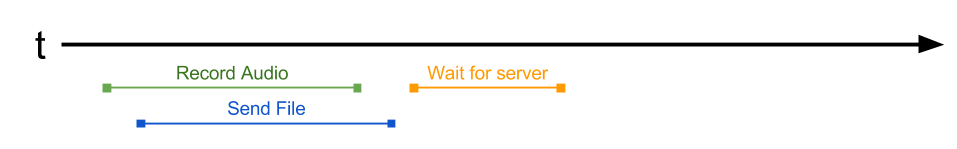}
	}
	
	\caption{The time required for speech recogntion by (a) sending the whole file, (b) streaming.}
	\label{fig:stream}
\end{figure}

Recording the full audio file, sending it to the Speech to Text service and getting back the audio transcription inevitably introduces some latency. In order to cope with this we adopt two strategies. The first solution is to minimize the latency itself. The IBM Speech to Text service allow to either send a whole audio file or to stream it. By starting to stream the audio file as soon as it's being recorded we are able to cut some of the latency introduced by the connection to a remote server. Fig~\ref{fig:stream} pictorially demonstrate how concurrently recording and streaming the audio file allows us to save some time. The second solution we adopt is to make this process transparent to the user. As soon as the robot stops recording we display on the chest tablet a fixed web page showing a loading gif and saying: "Processing audio input". Although this doesn't directly cut the latency it makes the users aware of the fact that the robot has heard their input and is currently processing it.

Once the server return the transcriptions of the audio recording the robot process the string locally and acts accordingly.
the user
\subsection{Face Recognition}
As mentioned in Section~\ref{sec:intro} we use the front camera positioned in the forehead for face recognition. The camera has a native resolution of $640*680$ and, although the NAOqi APIs present the ability to take pictures at higher resolution up to $1280*960$, we used the native resolution. The main reason behind our choice was that, in our testing taking picture at higher resolution considerably increased the amount of time needed for the robot to return the image (up to 5-6 sec). Similarly to what was done for speech recognition we set up a web service for face recognition. This service is not part of the IBM Bluemix Cloud service but was instead set up in-house.

The face recognition service, based on Deep Neural Network, underwent an initial training phase where we extracted meaningful features form thousands of images. At run time it offered two different functions: an enroll function and a query function. The enroll function requires an image and a label (the name or any identifier of the person in the picture). This function add the image to the gallery of faces the robot is able to recognize. The query function takes as input only a picture and returns, for each of the image currently enrolled in the gallery, a confidence value. 

When Pepper needs to run face recognition it simply take a picture and connects to the web service using the query function. It's also worth mentioning that both the enroll and query calls return an error if no face is visible in the picture while they select the biggest face if more then one is available. The choice of making the biggest face in the picture the center of the recognition was based on the assumption that the robot user is likely to be on the foreground.

\subsection{Tablet Integration}
Our goal is to have Pepper interact in the most natural way with its user. Although this would be possible by just using natural language the tablet offers a great opportunity to make the robot more transparent to the user and to overcome some of the traditional limitation of speech recognizer. In particular, our use of the chest tablet was aimed at: 1) improve the robot communication, 2) make the user aware of the internal processing of the robot, and 3) complement the speech recognition service in case of failure.

In order to improve the robot communication we use the tablet as a caption device that is paired with the robot text to speech. Whenever the robot says something the same text appears on the tablet and stays there until an other utterance follow or 10 seconds have passed; this is done by loading a dynamic web page on the chest tablet. By doing this we allow the user to follow the robot even in noisy environment or in case a distraction happens.

As mentioned in Section~\ref{sec:speechreco} the tablet also shows a static page with a loading gif and the text "Processing Audio Input" once the robot finished recording audio and it is waiting for the speech recognition result from the ASR server.

\begin{figure}[!ht]
	\centering
	\includegraphics[scale=0.24]{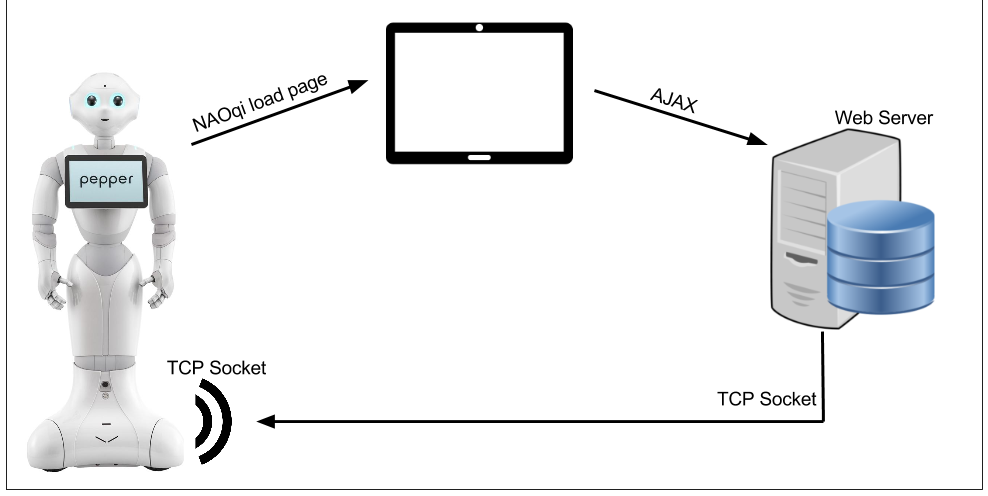}
	\caption{The connection between Pepper and its tablet. A page, with a text field is loaded on the tablet. The web server reads the input using an AJAX script and sends it back via TCP connections.}
	\label{fig:webserver}
\end{figure}

Finally to overcome the limitation of speech recognition, and in particular the problem of out-of-vocabulary words which is particularly significant with foreign proper names, we use the tablet as an additional input device that users can use to type. The NAOqi framework allows the tablet to load a web page and if a text field is present the Android interface automatically brings up a virtual keyboard on the screen. On the other hand NAOqi does not allow to directly read the input from the virtual keyboard. The solution we designed, shown in Figure~\ref{fig:webserver}, is based on three steps:
\begin{enumerate}
	\item When Pepper requires the user to type some input the specific web page is loaded on the tablet. This web page is hosted on a local web server.
	\item The robot open a TCP socket that the web serve is going to use as back-channel to return the input.
	\item The web server waits for user confirmation (via a button), records the input using an AJAX scripts, and sends the input back to the robot over TCP socket.
\end{enumerate}

%% file: results.tex
\section{Demonstration}

To demonstrate how the components described together can be used to enable Pepper to have an effective interaction with its user in this section we describe the demo we run for a group of middle-schooler visiting our lab. 

At the beginning of the demo Pepper is set to answer the commands only if the user speaking is recognized as one of the users already enrolled in the gallery of trusted users. The first users approaches the robot, starts the interaction by pressing the robot hand and asks for a hug. In the first attempt the users is correctly recognized by the face recognition service but the speech understanding, implemented as keyword search fails and Pepper ask the user to rephrase his request. In the two attempts the robot correctly recognize the user, understand his requests and behaves accordingly. Next a different user ask the robot for a hug. This time the user is not enrolled as a trusted user, is not recognized and therefore the robot refuses to comply with the request. Finally the initial user request to the robot to add a new person to the list of trusted users. After entering its name via the tablet and having a picture taken the second user is also able to have the robot perform task for him. The user can also give commands to move the robot.

%% file: conclusion.tex
\section{CONCLUSIONS}

This paper described our efforts to improve the capabilities of Pepper, adding personalized interactions to humans based on face recognition. We also used the tablet and LEDs to include non-verbal communication, and contributed a technique to simultaneous record and stream audio to a speech recognition web service in order to reduce the delay in the speech interaction. 

Furthermore, we showed how to include localization and navigation capabilities on Pepper using the ROS middleware. To the best of our knowledge, this is the
first work that combines ROS Indigo with the Pepper NAOqi framework. We provided instruction on how to install locally on Pepper and develop software for this integrated software architecture. 

Finally, as proof of concept, we show an "ownership" scenario where only authorized users can control the robot. We use face recognition to detect which user was giving commands to the robot, only responding if the user was the robot master. The commands the user can give include motion control, which use the ROS framework to make the robot navigate to different parts of the environment while localizing itself on a pre-build map of the world.

%


\bibliographystyle{IEEEtran}
\bibliography{main}

%% file: pepperMain.bbl
\begin{thebibliography}{1}
\providecommand{\url}[1]{#1}
\csname url@rmstyle\endcsname
\providecommand{\newblock}{\relax}
\providecommand{\bibinfo}[2]{#2}
\providecommand\BIBentrySTDinterwordspacing{\spaceskip=0pt\relax}
\providecommand\BIBentryALTinterwordstretchfactor{4}
\providecommand\BIBentryALTinterwordspacing{\spaceskip=\fontdimen2\font plus
\BIBentryALTinterwordstretchfactor\fontdimen3\font minus
  \fontdimen4\font\relax}
\providecommand\BIBforeignlanguage[2]{{%
\expandafter\ifx\csname l@#1\endcsname\relax
\typeout{** WARNING: IEEEtran.bst: No hyphenation pattern has been}%
\typeout{** loaded for the language `#1'. Using the pattern for}%
\typeout{** the default language instead.}%
\else
\language=\csname l@#1\endcsname
\fi
#2}}

\bibitem{ros2nao}
L.~L. Forero, J.~M. Y{\'a}nez, and J.~Ruiz-del Solar, ``Integration of the ros
  framework in soccer robotics: the nao case,'' in \emph{Robot Soccer World
  Cup}.\hskip 1em plus 0.5em minus 0.4em\relax Springer, 2013, pp. 664--671.

\bibitem{amcl}
D.~Fox, ``Adapting the sample size in particle filters through kld-sampling,''
  \emph{The international Journal of robotics research}, vol.~22, no.~12, pp.
  985--1003, 2003.

\bibitem{gmapping}
G.~Grisetti, C.~Stachniss, and W.~Burgard, ``Improving grid-based slam with
  rao-blackwellized particle filters by adaptive proposals and selective
  resampling,'' in \emph{Proceedings of the 2005 IEEE International Conference
  on Robotics and Automation}.\hskip 1em plus 0.5em minus 0.4em\relax IEEE,
  2005, pp. 2432--2437.

\bibitem{pepper-ros}
``{Pepper-Robot ROS Stack},'' \url{http://wiki.ros.org/pepper_robot}.

\bibitem{crosscomp}
``{Cross-compiling ROS for NAOqi},''
  \url{http://wiki.ros.org/nao/Installation/compileWithToolchain}.

\bibitem{naoqi}
``{NAOqi API and documentation},'' \url{http://doc.aldebaran.com/2-4/}.

\bibitem{ibm-speech}
``{IBM Speech to Text webservice},''
  \url{https://www.ibm.com/watson/developercloud/speech-to-text.html}.

\end{thebibliography}
